\documentclass[sigconf]{article}
\usepackage{ijcai25}

\usepackage{times}
\usepackage{soul}
\usepackage{url}
\usepackage[hidelinks]{hyperref}
\usepackage[utf8]{inputenc}
\usepackage[small]{caption}
\usepackage[switch]{lineno}
\usepackage{indentfirst}
\usepackage{graphicx}
\usepackage{multirow}
\usepackage{amsmath}
\usepackage{booktabs}
\usepackage{amsmath}
\usepackage{algorithm}
\usepackage{algpseudocode}
\usepackage{amsfonts}
\usepackage{mathtools}
\usepackage{float}
\usepackage{textcomp} 
\usepackage{placeins}
\usepackage{tabularx}
\usepackage{adjustbox}

\title{Active Learning from Scene Embeddings for End-to-End Autonomous Driving}

\author{
Wenhao Jiang$^1$
\and
Duo Li$^2$\and
Menghan Hu$^{1}$ \and
Chao Ma$^3$ \and
Ke Wang$^2$ \and
Zhipeng Zhang$^2$ \\
\affiliations
$^1$Eact China Normal University, 
$^2$KargoBot,
$^3$Shanghai Jiao Tong University \\
\emails
51255904051@stu.ecnu.edu.cn,
\{liduo, kewang\}@kargobot.ai,
mhhu@ce.ecnu.edu.cn,
chaoma@sjtu.edu.cn,
zhipeng.zhang.cv@outlook.com
}

\begin{document}

\maketitle
\begin{abstract}
\indent In the field of autonomous driving, end-to-end deep learning models show great potential by learning driving decisions directly from sensor data. However, training these models requires large amounts of labeled data, which is time-consuming and expensive. Considering that the real-world driving data exhibits a long-tailed distribution where simple scenarios constitute a majority part of the data, we are thus inspired to identify the most challenging scenarios within it. Subsequently, we can efficiently improve the performance of the model by training with the selected data of the highest value. Prior research has focused on the selection of valuable data by empirically designed strategies. However, manually designed methods suffer from being less generalizable to new data distributions. Observing that the BEV (Bird’s Eye View) features in end-to-end models contain all the information required to represent the scenario, we propose an active learning framework that relies on these vectorized scene-level features, called $SEAD$. The framework selects initial data based on driving-environmental information and incremental data based on BEV features. Experiments show that we only need 30\% of the nuScenes training data to achieve performance close to what can be achieved with the full dataset. Source code will be released.
\end{abstract}
\section{Introduction}
\label{sec:intro}

\indent In recent years, deep learning has made breakthroughs across various fields \cite{he2016deep,vaswani2017attention,zhu2020deformable}, with autonomous driving (AD) being one of the most promising areas of application. Early automated driving systems were primarily based on modular architectures, which relied on several independent subsystems to perform tasks such as perception, prediction, decision-making, and control \cite{behere2015functional,xu2021autonomous}. While this approach has been successful to a certain extent, the subsystems need to manually design features and perform complex parameter tuning, which leads to high system development and maintenance costs and difficulty in handling complex and changing driving environments.

To overcome these limitations, end-to-end AD (E2E-AD) algorithms have gradually become a research hotspot \cite{hu2022st,hu2023planning,jiang2023vad}. E2E-AD integrates modules such as perceptual prediction planning into a pipeline, taking raw sensor data as input and directly outputting driving behavior. It realizes the fully differentiable learning of the whole process, eliminates the need for manual feature engineering and modular pipeline, and makes the system architecture more concise and efficient.

In complex and dynamic driving environments, E2E approaches demonstrate the potential to handle diverse scenarios and better tackle real-world challenges. However, E2E-AD models have higher data requirements \cite{caesar2020nuscenes,jia2024bench2drive}. Training such models requires substantial scene data with detailed labeling information (e.g., 3D bounding boxes, segmentation masks), making data preparation a time-intensive and costly endeavor. To alleviate these labeling costs, active learning has emerged as a promising solution. By selecting the most informative samples for labeling, active learning aims to achieve high model performance with fewer labeled examples, thus reducing both labeling and training expenses \cite{ren2021survey,zhan2022comparative}.

Currently, some studies are beginning to explore active learning methods in E2E-AD algorithms. For instance, \cite{lu2024activead} introduces data selection strategies tailored to end-to-end scenarios with a focus on planning. However, their approach relies on the intermediate outputs of the model, e.g., the model's perceptual ability, and action prediction ability. Some E2E models lack the computation of these intermediate quantities, which limits their ability to use this method. Besides, their scheme is artificially crafted and its robustness needs further experimental validation. Additionally, their data selection remains at the scene level, without exploring the feasibility of using more refined consecutive key frames. 

There is an urgent need to find a data selection strategy that does not depend on manually set selection rules and is more generalizable to different data distributions. BEV algorithms \cite{philion2020lift,huang2021bevdet,liu2023bevfusion} shine in AD by virtue of their ability to comprehensively perceive the environment, efficiently fuse multi-sensor data, and improve system stability and reliability. BevFormer \cite{li2022bevformer} achieved state-of-the-art performance in perception tasks using solely visual inputs. As the latest E2E-AD models increasingly rely on BEV features, filtering data within AD datasets based on BEV features presents a promising direction for future advancements.

Inspired by the above phenomenon, we propose an active learning method based on BEV features for E2E-AD. The BEV features are used to extract the scene-level embedding. Our method aligns with the standard active learning framework: initial training, data filtering, and retraining. 
We first leverage the abundant dynamic and static information in AD datasets to construct a diverse initial dataset, which is used to train a baseline model. Using this baseline model, we extract BEV features from the remaining data to evaluate scene value based on the predefined rule. The rule can be summarized as extracting key elements from BEV features and calculating the cumulative changes in these elements at the clip level as the selection criterion. For valuable scenes, key consecutive frames are selected and labeled, then combined with the existing data for further rounds of training and refinement. 

We rigorously designed and conducted fair comparative experiments to validate the effectiveness of our proposed approach. The experimental results demonstrated a notable improvement in model planning performance, achieved under identical or even reduced labeling budgets. We further performed ablation studies to systematically examine the contribution of each individual module within our approach. Additionally, we included visualizations of selected frames to provide an intuitive understanding of the internal mechanics and impact of these modules. To comprehensively evaluate the robustness of our algorithm, we extended its application to multiple datasets, enabling us to verify its adaptability and stability across diverse data scenarios. Our principal contributions can be summarized as follows:

\begin{itemize}
    \item We prioritized the selection of data with high utility for End-to-End Autonomous Driving (E2E-AD) from a Bird's Eye View (BEV) perspective. This targeted data selection is instrumental in optimizing the performance of the E2E-AD system.

    \item Our approach achieved competitive performance while significantly reducing labeling requirements, demonstrating not only its efficiency but also its potential to address challenges in practical applications where labeling resources are limited or costly. 

    \item We successfully extended our algorithm to a broader spectrum of models and datasets. This extension substantiates the robustness of our algorithm, demonstrating its capacity to deliver stable and reliable performance across various environments and frameworks.
\end{itemize}
\section{Related Work}
\label{sec:relatedwork}

\subsection{End-to-End Autonomous Driving}
E2E-AD aims to cope with the risks of information loss, error accumulation, and feature misalignment that may result from the separation of traditional model modules \cite{chen2024end,tampuu2020survey,hu2023planning}. Current algorithms mainly follow the idea of imitation learning, i.e., training a model by mimicking the behavior of an expert. CILRS \cite{codevilla2019exploring} investigates the advantages and limitations of imitation learning in E2E-AD and gives new benchmarks. LBC \cite{chen2020learning} uses an agent model that has access to privileged information to guide the training of a purely visually based sensorimotor agent. Transfuser \cite{prakash2021multi} and InterFuser \cite{shao2023safety} propose the use of a sensor fusion Transformer, which combines the benefits of visual imagery and LiDAR to achieve advanced driving performance. ST-P3 \cite{hu2022st} uses a spatial-temporal feature learning scheme to achieve performance gains with purely vision-based inputs. UniAD \cite{hu2023planning} designs an end-to-end autopilot framework that integrates full-stack tasks in a planning-oriented manner, jointly optimizing intermediate modules. VAD \cite{jiang2023vad} uses vectorized scenario representations to improve model inference speed significantly. Some works use reinforcement learning ideas to train models \cite{zhang2022rethinking,kiran2021deep}. \cite{toromanoff2020end,chekroun2023gri} use supervised learning to obtain a scene representation on which a shallow policy head is trained. \cite{knox2023reward,zhang2021end} design a reward function based on prior access to privileged information.
\subsection{Active Learning}
Active learning is a well-established approach that aims to minimize the annotation cost by selectively querying the most informative data points for labeling \cite{ren2021survey,zhan2022comparative}. Active learning algorithms primarily focus on uncertainty \cite{lewis1994heterogeneous,gao2020consistency,joshi2009multi} and data diversity \cite{sinha2019variational,agarwal2020contextual,xie2023active}. Uncertainty-based methods prioritize the selection of samples with the highest uncertainty in model predictions, aiming to enhance the model’s generalization capability. Uncertainty can be measured through posterior probabilities \cite{lewis1994heterogeneous}, margin sampling \cite{roth2006margin}, or entropy \cite{joshi2009multi}. Some studies leverage Bayesian inference, using techniques such as Monte Carlo Dropout \cite{gal2016dropout} or Deep Gaussian Processes \cite{damianou2015deep} to estimate predictive variance through multiple forward passes.
Diversity-based strategies aim to select subsets of unlabeled data that comprehensively cover the data space, minimizing redundancy and enhancing generalization. Representative methods include clustering-based selection, core-set strategies \cite{sener2017active} that minimize the maximum distance between selected samples and the rest of the dataset, and submodular optimization \cite{elhamifar2013convex} to select samples with the highest marginal gains. In addition to these, some algorithms assess the influence of data on the model \cite{liu2021influence,chhabra2024data}. ISAL\cite{liu2021influence} proposes an approach to estimate the expected gradient of unlabeled samples, quantifying their potential impact on model performance and guiding the selection process accordingly. \cite{lu2024activead} first combines active learning with E2E-AD, though their strategy's generalizability requires further validation. Our approach offers a simple and efficient integration into BEV-based AD models.
\section{Methods}

\begin{figure*}[!h]
\centering
\includegraphics[width=\textwidth]{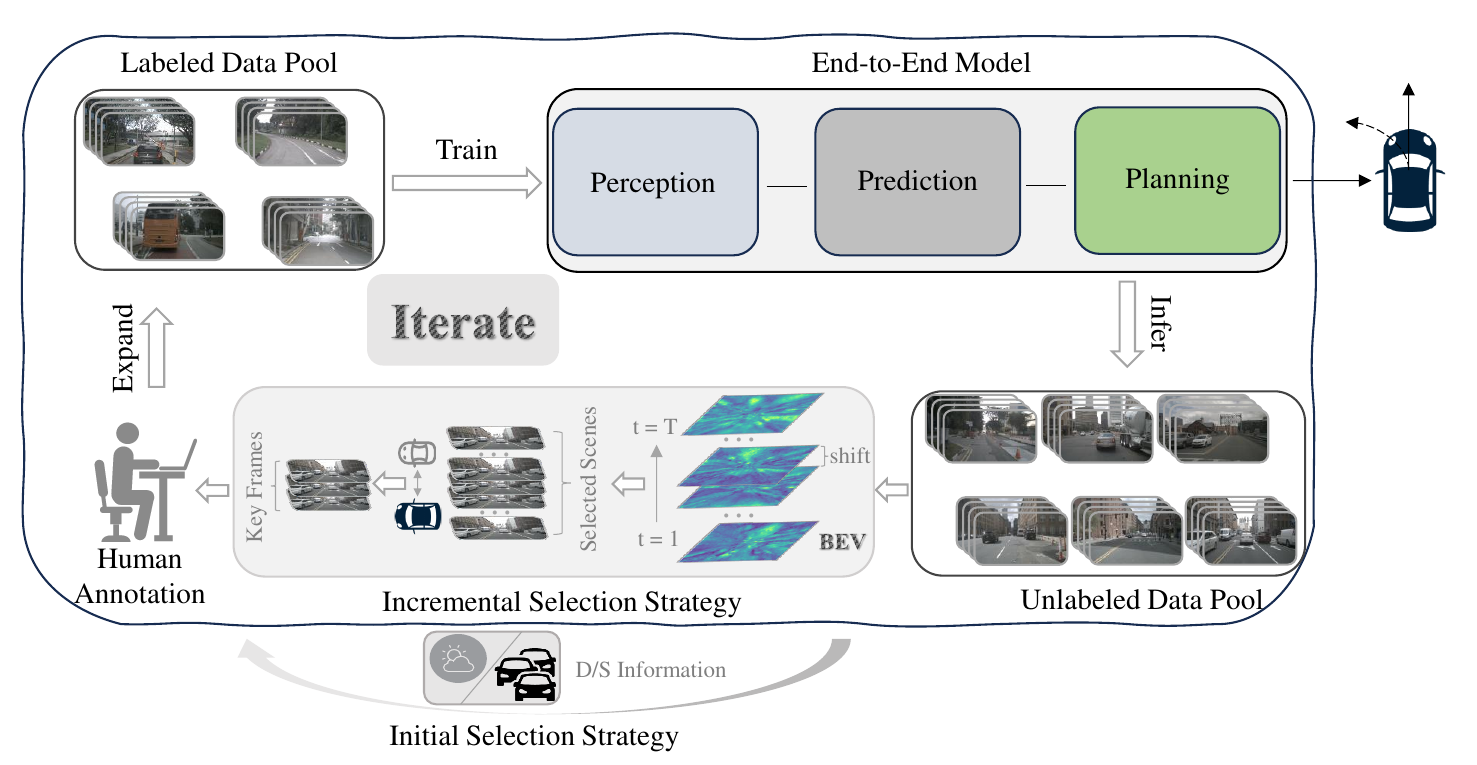} 
\caption{\textbf{Overall pipeline of SEAD}. Following the active learning setup, we first build an initial training set based on the initial strategy to train the model. Subsequently, we leverage the trained model and incremental selection strategy to actively select new data, iterating this process continuously. In this context, $D/S$ represents dynamic/static.}
\label{fig:1}
\end{figure*}

\subsection{Problem Definition}
Our optimization objective for applying active learning to E2E-AD tasks is defined as \( G = Max(\frac{P}{B})\), where \( P \) represents model planning performance and \( B \) denotes the labeling budget. This introduces two optimization strategies:
\begin{enumerate}
    \item Given a target planning performance \( P \), reduce the labeling budget \( B \);
    \item Given a labeling budget \( B \), improve the planning performance \( P \).
\end{enumerate}
In this paper, we adopt the second strategy.

Currently, commonly used AD datasets are structured at the clip-level. Specifically, the training set of a comprehensive AD dataset consists of $N$ clips, each capturing a continuous driving scene over $T$ frames.  Within each clip, frames maintain contextual continuity with one another, making the use of clips as fundamental training units beneficial for tasks such as object tracking and route planning. Empirical evidence, as well as analysis of the data itself, reveals that corner cases represent only a small portion of all $N$ clips. Furthermore, each clip may contain timestamped samples corresponding to repeated, simple scenarios. 

We propose an active learning strategy designed to identify and extract valuable clips from the data pool, followed by the selection of key frames. Initially, given an available unlabeled data pool $P = \{x_i\}_{i \in [N]}$, we leverage the prior information within AD datasets to select valuable clip indices $\mathcal{K}$ and construct the initial training set $\hat{P} = \{x_i, y_i\}_{i \in [\mathcal{K}]}$. The model \(f \) is trained on $\hat{P}$. For incremental selection, \(f \) performs inference on the remaining data pool, with key frames selected according to predefined criteria. Combine selected key frames with $\hat{P}$ for further training to update \(f \). This incremental process is repeated until the total data reaches the budget \( B \), depicted in Fig. \ref{fig:1}.

\subsection{Initial Dataset Construction Based on Dynamic and Static Information}
\begin{figure}[htbp]
\centering
\includegraphics[width=\columnwidth]{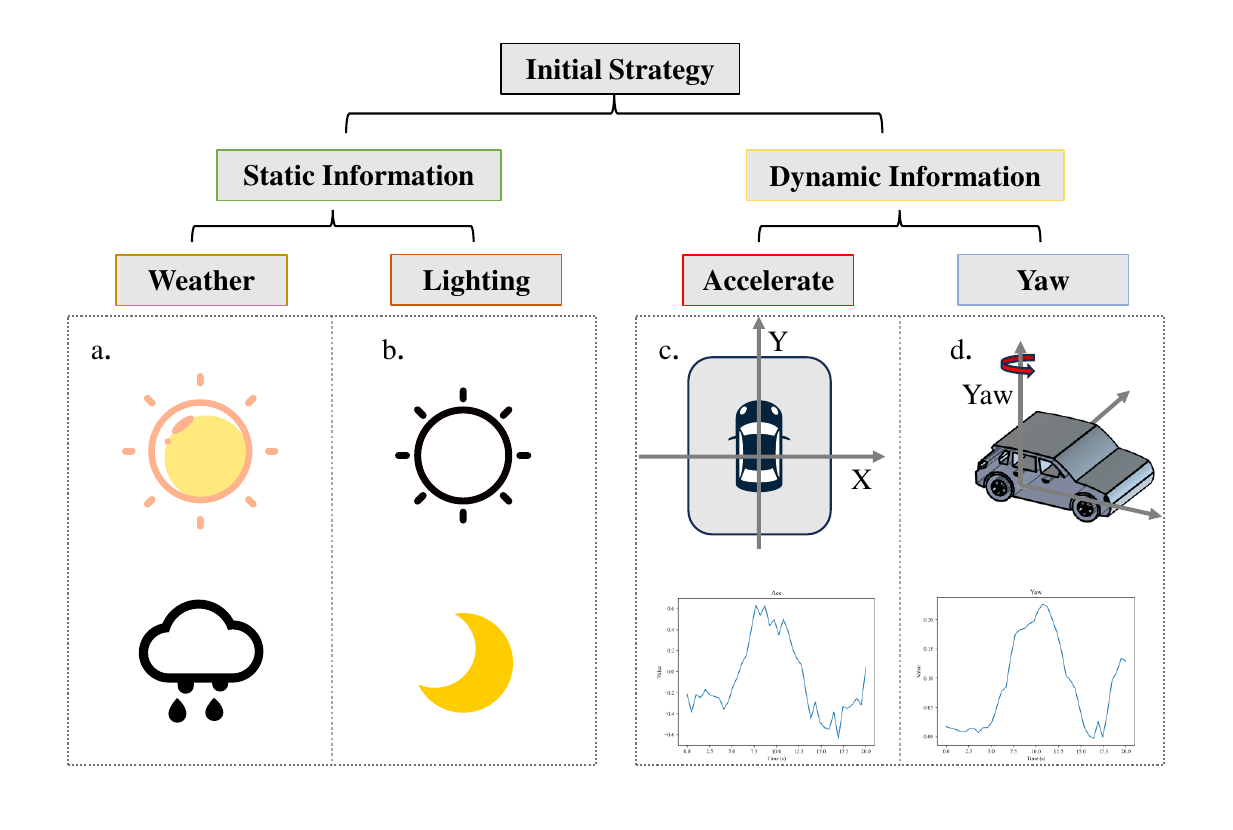} 
\caption{\textbf{Initial selection}. Constructing a diverse initial dataset by leveraging the rich information within the AD dataset.}
\label{fig:2}
\end{figure}

In conventional active learning algorithms for general tasks (e.g., classification, object detection), a randomly selected subset of data is typically used to train the initial model. This approach is feasible because the data and labels in these tasks are relatively straightforward and lack additional contextual information. However, in autonomous driving (AD) tasks, a wealth of additional information is collected by the vehicle's sensors, such as lighting conditions, weather, speed, and cornering angle. This auxiliary information is readily available without extra labeling effort, allowing us to leverage it to construct an initial dataset that encompasses diverse scenarios. Following the structure of the nuScenes dataset \cite{caesar2020nuscenes}, we build our initial dataset along two dimensions: static scene information and dynamic vehicle information, as illustrated in Fig. \ref{fig:2}. Specifically, for static scene information, inspired by \cite{lu2024activead}, we categorize the training set into four subsets \{DS, DR, NS, NR\} based on a combination of lighting conditions (day or night) and weather conditions (sunny or rainy). 

Building on this, we further enhance the diversity by incorporating dynamic information gathered during vehicle motion. Information on vehicle acceleration and turning angle is available in the CAN bus data, from which we select the vehicle’s $y$-axis acceleration and yaw deflection.   

Using these two parameters, we compute the standard deviation of acceleration and the cumulative turning angles within each scene to capture the dynamics of vehicle motion. After sorting based on these metrics, we obtain $Acc$ and $Yaw$ scores and apply a selection ratio, denoted as $\lambda$, to categorize the data into ``valuable" and ``normal" subsets. This results in four subsets: $Acc_v$, $Acc_n$, $Yaw_v$, and $Yaw_n$. The training set $D_T$ is further divided into three subsets based on scene composition: scenes containing both $Acc_v$ and $Yaw_v$, scenes containing only one, and scenes with neither.

\begin{equation}
\label{eq:1}
\begin{split}
    B = Acc_v \cap Yaw_v\\
    N = Acc_n \cap Yaw_n\\
    S = D_T - (B \cap N)
\end{split}
\end{equation}

Based on the static scene information and dynamic vehicle information, we can divide the whole training set into 12 subsets \{DS-B, DS-S, DS-N, ..., NR-B, NR-S, NR-N\}\ containing different driving behavior data in different scenes.

\subsection{Key Scene-to-Frames based on BEV Features}
In the initial selection phase, both static environmental and dynamic driving data were utilized. For incremental selection, the trained model was then used to extract dynamic and static information from the remaining data pool. We define dynamic information as the cumulative changes between frames throughout an entire scene, and static information as the interactions between scene participants within a specific frame. Based on these two types of information, we implemented a layered filtering process, progressing from the data pool to valuable scenes and subsequently to key frames.

\subsubsection{Select Valuable Scenes Based on Dynamic Information}
The model trained in the previous stage takes multi-view RGB images as input, and through a feature extraction module and a BEV encoder module, generates BEV features.
A straightforward approach to measure changes between frames is to compute the feature shifts \( FS \) between two BEV representations. However, the high dimensionality of the BEV feature tensor makes direct computation time-intensive. Additionally, the richness of information within BEV feature maps could obscure the truly valuable parts. Given that the primary elements in road scenes are the map and agents, we can transform the task of computing \( FS \) between BEV frames into calculating the shifts in map and agent features specifically. This approach not only reduces computational demands but also concentrates on the information that is most relevant.

Taking agents as an example, the process is as follows: a query vector \( Q_a \) is constructed for the agent, with the previously obtained BEV features used as the value. Following the approach \cite{zhu2020deformable}, this yields the agent's feature representation \( F_a \).
In the clip-level data scenario, each scene contains  $T$ instances of
 \( F_a \). The Earth Mover’s Distance (EMD) \cite{rubner2000earth} is employed to measure the difference between two consecutive:
\begin{equation}
\label{eq:2}
FS = EMD(F_{cur}, F_{prev})
\end{equation}

\begin{figure}[tbp]
\centering
\includegraphics[width=\columnwidth]{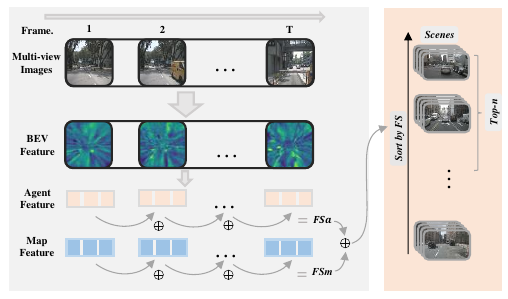} 
\caption{\textbf{Calculate $FS$ for scenes}. Convert the computation of BEV $FS$ to focus on the shifts of key elements, specifically the agent and map features. Then, accumulate the frame-to-frame $FS$ results to determine the $FS$ for the scene.}
\label{fig:3}
\end{figure}

When the dimensionality of \( F_a \) is high, directly calculating the \( FS_a \) between two \( F_a \) becomes computationally expensive and resource-intensive. To reduce the computational complexity, we first apply a dimensionality reduction strategy to project \( F_a \) into a lower-dimensional space, denoted as \( F_a' \), and then compute the \( FS_a \) between \( F_a' \). The calculation process for the map is analogous, resulting in \( FS_m \). The final \( FS_s \) of the scene is calculated as:
\begin{equation}
\label{eq:3}
FS_s = \sum_{t=1}^T (FS_{a,t} + FS_{m,t})
\end{equation}
We select the top \( n \) clips with the largest \( FS_s \) from the data pool, where \( n \) represents the phase budget.
The detailed process is illustrated in Fig. ~\ref{fig:3}.

\subsubsection{Select Key Frames Based on Static Information}
When the distance between the ego vehicle and other road participants is large, the driving difficulty tends to be lower. From the perspective of interaction, we extract key frames from a single clip. Assume a clip contains \( T \) frames, represented as \( C_i = [f_1, f_2, \dots, f_T] \). For a specific frame \( f_k \), after obtaining the queried feature \( F_a \), a regression network is used for inference to provide predictions about the agent, including class, position, confidence, and other relevant information. By setting a confidence threshold \( \theta_{\tau} \), targets with low confidence scores are filtered out, leaving the remaining targets with their corresponding positions \( P = [p_1, p_2, \dots, p_n] \). The minimum distance between these remaining targets and the ego vehicle is then computed, and if this distance is smaller than the predefined distance threshold \( \theta_d \), the current frame is identified as a key frame:
\begin{equation}
\label{eq:4}
y_k =
\begin{cases} 
1 & \text{if } \min\limits_{i\in [1, n]} (p_{\text{ego}} - p_i) < \theta_d, \\
0 & \text{otherwise.}

\end{cases}
\end{equation}
For each clip, we obtain an array \( K = [y_1, y_2, \dots, y_T] \) of length \(T \), which indicates whether each frame is a key frame. To ensure temporal continuity in the driving scenario, we select the longest sequence of consecutive key frames for subsequent annotation and training.

The workflow of the entire program is outlined in Alg. \ref{alg:1}, covering both the initial selection and subsequent incremental selections. Given a total budget \( B \), data samples are selected in batches of size \( n_{itr} \) at each iteration. The model is iteratively trained on all previously selected data, followed by a new data selection round, until the budget \( B \) is exhausted. 
\section{Experiments}

\subsection{Experimental Setup}
We conduct experiments on nuScenes \cite{caesar2020nuscenes}, a commonly used benchmark dataset for AD, with experimental settings aligned to previous work UniAD \cite{hu2023planning}, VAD \cite{jiang2023vad}.
\subsubsection{Data/Metrics}
The nuScenes dataset comprises a total of 1,000 diverse driving scenarios: 700 for training, 150 for validation, and 150 for testing, with each scenario lasting about 20 seconds. Data collection uses a vehicle outfitted with 6 cameras, 1 LiDAR, and 5 radars, offering a 360° horizontal field of view. Keyframes are annotated at a 2 Hz sampling rate, capturing comprehensive environmental details and vehicle driving behavior. We continue the setting of ActiveAD \cite{lu2024activead} by setting a 30\% annotation budget and selecting 10\% of the data in each of the three batches.

The evaluation metrics employed in this study are aligned with the open-loop metrics of previous work \cite{hu2022st,hu2023planning,jiang2023vad}, and displacement error (L2) and collision rate (CR) are chosen to measure the planning performance of the model.

\begin{algorithm}[htbp]
\caption{Pseudo-code for SEAD}
\label{alg:1}
\begin{algorithmic}[1]
\Require Unlabeled pool $P = \{x_i\}_{i \in [N]}$, labeled pool $ \hat{P} = \emptyset$, annotation budget $B$, iteration $I$, model $f_{itr}$,  iterated selection $n_{itr}$
\State Selected clip index $\mathcal{K} = \emptyset$  
\For {$itr$ in range($I$)}
\If{itr == 0}
    \State Calculate nums $ \hat{n}_{x,y} = n_{itr}\times n_{x,y} / \sum_{x,y} n_{x,y}$, where $x \in \{\text{DS, DR, NS, NR}\}, y \in \{\text{B, S, N}\}$
    \State Random select $\hat{n}_{x,y}$ samples in $P_{x,y}$ and add their indexes to $\mathcal{K}$.
    \State Update $\hat{P} = \{x_i, y_i\}_{i \in [\mathcal{K}]}$
\Else
    \State Update $P_{itr} = P - P_\mathcal{K}$.
    \State Inference on $P_{itr}$ with the model $f_{itr-1}$ to calculate 
    $FS$ for each clip with Eq. \ref{eq:2},\ref{eq:3}.
    \State Sort the clip in $P$ in descending order of $FS$
    \State Select the former $n_{\text{itr}}$ indexes and add to $\mathcal{K}$
    \State Select key frames index $k$ for each selected clip with Eq. \ref{eq:4} 
    \State Update $\hat{P} = \hat{P} + \{x_{i,k}, y_{i,k}\}_{i \in [\mathcal{K}]}$
\EndIf
\State Train the model $f_{itr}$ from scratch using $\hat{P}$.
\EndFor
\end{algorithmic}
\end{algorithm}

\subsubsection{Models/Training}
We select VAD as our base model. Compared to previous works such as ST-P3 and UniAD, VAD leverages a vectorized scene representation to enable faster inference and improved open-loop performance. Additionally, the lightweight version, VAD-Tiny, requires less training time than the standard VAD-Base, with only minimal performance degradation. Therefore, we adopt VAD-Tiny for rapid solution validation.

We use AdamW \cite{loshchilov2017decoupled} optimizer and Cosine Annealing \cite{loshchilovstochastic} scheduler to train VAD-Tiny with weight decay 0.01 and initial learning rate $2\times10^{-4}$. The approach includes the following hyper parameters: selection ratio  $\lambda$ set to 1/3, confidence threshold \( \theta_{\tau} \) set to 0.5, and distance threshold \( \theta_d \) set to 5.0 m. We reduce EMD computation complexity by averaging the elements. VAD-Tiny is trained for 20 epochs on 8 NVIDIA GeForce RTX 4090 GPUs with batch size 1 per GPU.

\begin{table*}[!h]
\centering
\caption{Planning Performance. Comparative experimental results are referenced from \protect\cite{lu2024activead}.}
\label{table:1}
\small
\begin{tabular*}{\textwidth}{@{\extracolsep{\fill}} c|c|c|cccc|cccc}
\toprule
\multicolumn{1}{c|}{\multirow{2}{*}{\textbf{Base Model}}} &
\multicolumn{1}{c|}{\multirow{2}{*}{\textbf{Percent}}} &
\multicolumn{1}{c|}{\multirow{2}{*}{\textbf{Method}}} &
\multicolumn{4}{c|}{\textbf{L2 (m)} $\downarrow$} &
\multicolumn{4}{c}{\textbf{CR (\%)} $\downarrow$} \\
\cmidrule(lr){4-7} \cmidrule(lr){8-11}
& & & 1s & 2s & 3s & Avg. & 1s & 2s & 3s & Avg. \\
\midrule
ST-P3 & 100\% & * & 1.33 & 2.11 & 2.90 & 2.11 & 0.23 & 0.62 & 1.27 & 0.71 \\
UniAD & 100\% & * & 0.42 & 0.64 & 0.91 & 0.67 & * & * & * & * \\
VAD-Base & 100\% & * & 0.39 & 0.66 & 1.01 & 0.69 & 0.08 & 0.16 & 0.37 & 0.20 \\
VAD-Tiny & 100\% & * & 0.38 & 0.68 & 1.04 & 0.70 & 0.15 & 0.22 & 0.39 & 0.25 \\
\midrule
\multirow{4}{*}{VAD-Tiny} & 10\% & Random & 0.51 & 0.83 & 1.23 & 0.86 & 0.40 & 0.62 & 0.98 & 0.67 \\
& 10\% & ActiveFT & 0.54 & 0.88 & 1.29 & 0.90 & 0.20 & 0.41 & 0.81 & 0.47 \\
& 10\% & ActiveAD & \textbf{0.47} & 0.80 & 1.21 & 0.83 & 0.13 & 0.35 & 0.80 & 0.43 \\
& 10\% & SEAD(Ours) & 0.50 & \textbf{0.80} & \textbf{1.17} & \textbf{0.82} & \textbf{0.13} & \textbf{0.25} & \textbf{0.53} & \textbf{0.30} \\
\midrule
\multirow{6}{*}{VAD-Tiny} & 20\% & Random & 0.49 & 0.80 & 1.17 & 0.82 & 0.36 & 0.49 & 0.77 & 0.54 \\
& 20\% & Coreset & 0.48 & 0.78 & 1.16 & 0.81 & 0.20 & 0.40 & 0.69 & 0.43 \\
& 20\% & VAAL & 0.54 & 0.89 & 1.31 & 0.91 & 0.17 & 0.38 & 0.66 & 0.40 \\
& 20\% & ActiveFT & 0.50 & 0.82 & 1.21 & 0.84 & 0.27 & 0.42 & 0.63 & 0.44 \\
& 20\% & ActiveAD & 0.44 & 0.73 & 1.10 & 0.76 & 0.18 & 0.36 & 0.62 & 0.39 \\
& 20\% & SEAD(Ours) & \textbf{0.44} & \textbf{0.71} & \textbf{1.04} & \textbf{0.73} & \textbf{0.12} & \textbf{0.24} & \textbf{0.47} & \textbf{0.28} \\
\midrule
\multirow{6}{*}{VAD-Tiny} & 30\% & Random & 0.45 & 0.76 & 1.12 & 0.78 & 0.17 & 0.30 & 0.63 & 0.37 \\
& 30\% & Coreset & 0.43 & 0.71 & 1.06 & 0.73 & 0.43 & 0.51 & 0.68 & 0.54 \\
& 30\% & VAAL & 0.46 & 0.79 & 1.19 & 0.81 & 0.18 & 0.33 & 0.54 & 0.35 \\
& 30\% & ActiveFT & 0.46 & 0.76 & 1.13 & 0.78 & 0.18 & 0.35 & 0.63 & 0.39 \\
& 30\% & ActiveAD & 0.41 & 0.66 & 0.97 & 0.68 & 0.10 & 0.18 & \textbf{0.36} & 0.21 \\
& 30\% & SEAD(Ours) & \textbf{0.33} & \textbf{0.59} & \textbf{0.92} & \textbf{0.61} & \textbf{0.04} & \textbf{0.18} & 0.42 & \textbf{0.21} \\
\bottomrule
\end{tabular*}
\end{table*}

\begin{table*}[htbp]
\centering
\caption{Performance comparison across different weather, lighting, and driving-command scenarios. Average L2 (m) / Average CR (\%) under a 30\% annotation budget are used as evaluation metrics.}
\label{table:3}
\resizebox{\textwidth}{!}{
\begin{tabular}{cccccccccc}
\toprule
\multirow{2}{*}{\textbf{Method}} & \multicolumn{4}{c}{\textbf{Weather / Lighting}} & \multicolumn{4}{c}{\textbf{Driving-Command}} & \multirow{2}{*}{\textbf{All}} \\
\cmidrule(lr){2-5} \cmidrule(lr){6-9}
 & Day & Night & Sunny & Rainy & Go Straight & Turn Left & Turn Right & Overtake & \\
\midrule
Complete & 0.67 / 0.27 & 1.01 / \textbf{0.14} & 0.70 / 0.32 & 0.72 / \textbf{0.04} & 0.69 / 0.32 & 0.74 / \textbf{0.13} & 0.67 / 0.20 & 0.84 / 0.13 & 0.70 / 0.25 \\
Random & 0.72 / 0.26 & 1.29 / 1.25 & 0.78 / 0.39 & 0.79 / 0.26 & 0.70 / 0.22 & 0.89 / 1.03 & 0.86 / 0.32 & 1.05 / 0.22 & 0.78 / 0.37 \\
Coreset & 0.71 / 0.57 & 0.97 / 0.27 & 0.72 / 0.65 & 0.78 / 0.06 & 0.69 / 0.67 & 0.78 / 0.31 & 0.78 / 0.38 & 0.96 / 0.14 & 0.73 / 0.54 \\
VAAL & 0.78 / 0.34 & 1.09 / 0.34 & 0.80 / 0.40 & 0.89 / 0.12 & 0.79 / 0.38 & 0.86 / 0.34 & 0.82 / 0.20 & 0.96 / 0.18 & 0.81 / 0.35 \\
ActiveFT & 0.76 / 0.37 & 1.08 / 0.43 & 0.79 / 0.40 & 0.78 / 0.28 & 0.70 / 0.35 & 0.88 / 0.62 & 0.91 / 0.20 & 1.18 / 0.44 & 0.79 / 0.38 \\
ActiveAD & 0.64 / 0.20 & 1.03 / 0.31 & 0.68 / 0.24 & 0.68 / 0.07 & 0.62 / \textbf{0.21} & 0.74 / 0.25 & 0.80 / 0.20 & 0.85 / 0.13 & 0.68 / 0.21 \\
SEAD(Ours) & \textbf{0.59 / 0.20} & \textbf{0.78} / 0.30 & \textbf{0.62 / 0.23} & \textbf{0.58} / 0.13 & \textbf{0.59} / 0.22 & \textbf{0.63} / 0.28 & \textbf{0.66 / 0.17} & \textbf{0.67 / 0.02} & \textbf{0.61 / 0.21} \\
\bottomrule
\end{tabular}}
\end{table*}

\subsection{Benchmark Comparison}
In Tab. \ref{table:1}, we compare the planning performance of various algorithms under annotation budgets of 10\%, 20\%, and 30\%, including ActiveFT \cite{xie2023active}, ActiveAD \cite{lu2024activead}, Coreset \cite{sener2017active}, and VAAL \cite{sinha2019variational}. It is worth noting that the selection ratios of 20\% and 30\% are calculated at the clip level. Since we adopt a consecutive keyframe strategy, the actual annotation volume is lower than these values. In In \cite{jia2024bench2drive,zhai2023rethinking}, the authors highlight the limitations of open-loop metrics, noting they may not reliably reflect real-world performance. Our experiments show a weak positive correlation between L2 and CR metrics, with improvements in one often accompanied by declines in the other. Therefore, when evaluating model planning performance, we consider both of these metrics comprehensively. From the table, we can see that with the VAD-Tiny model, our approach achieves performance comparable to that of the full dataset with just 30\% of the annotation budget. Compared to traditional active learning algorithms, our method also demonstrates a significant performance advantage. ActiveAD is an active learning solution designed for E2E-AD, which selects data based on meticulously crafted rules. In contrast, our BEV feature-based approach achieves superior planning performance with a lower annotation budget.

\subsection{Ablation Study}

\subsubsection{Planning performance on fine-grained validation set}
In the previous section, we noted that the AD dataset exhibits a long-tailed distribution. To mitigate the risk of overfitting the model to more common, straightforward scenarios, we subdivided the validation set based on specific driving behaviors and scenarios, following \cite{lu2024activead}. This allowed us to assess the model’s performance across several distinct scenario categories, with results summarized in Tab. \ref{table:3}. The results show that our approach does not overfit simple scenarios and achieves competitive performance in relatively complex scenes, such as nighttime, rainy conditions, turns, and overtaking maneuvers.

\begin{table*}[htbp]
\centering
\caption{Effectiveness of modules. “RA” and “DS” represent random selection and initial dataset construction based on dynamic/static information, respectively. “FS” and “KF” indicate the use of feature shifts and key frames, respectively.}
\label{table:4}
\resizebox{\textwidth}{!}{
\begin{tabular}{ccc|cc|ccc|cccc}
\toprule
\multirow{2}{*}{\textbf{ID}} & \multicolumn{2}{c|}{\textbf{Initiation}} & \multicolumn{2}{c|}{\textbf{Active Selection}} & \multicolumn{3}{c|}{\textbf{Average L2 (m)}} & \multicolumn{3}{c}{\textbf{Average CR (\%)}} \\
\cmidrule(lr){2-3} \cmidrule(lr){4-5} \cmidrule(lr){6-8} \cmidrule(lr){9-11}
 & RA & DS & FS & KF & 10\% & 20\% & 30\% & 10\% & 20\% & 30\% \\
\midrule
1 & \checkmark & - & - & - & 0.86 & 0.82 & 0.78 & 0.67 & 0.54 & 0.37 \\
2 & - & \checkmark & - & - & 0.82 (-0.04) & 0.65 (-0.17) & 0.67 (-0.11) & 0.30 (-0.24) & 0.33 (-0.04) & 0.23 (-0.14) \\
3 & - & \checkmark & \checkmark  & - & 0.82 (-0.04) & 0.72 (-0.10) & 0.65 (-0.13) & 0.37 (-0.17) & 0.32 (-0.22) & 0.31 (-0.06) \\
4 & - & \checkmark & \checkmark & \checkmark & 0.82 (-0.04) & 0.73 (-0.09) & 0.61 (-0.17) & 0.30 (-0.37) & 0.28 (-0.26) & 0.21 (-0.16) \\
\bottomrule
\end{tabular}}
\end{table*}

\subsubsection{Effectiveness of Modules}
To evaluate the effectiveness of each module, we conduct detailed ablation experiments to systematically assess their contributions. The experimental results in Tab. \ref{table:4}  reveal some interesting insights. The initial selection strategy, designed based on prior information, significantly outperforms random selection in terms of performance. Besides, more data doesn’t always yield better model performance; for instance, by using selected key frames instead of entire clips, our approach achieves better collision performance while reducing the data volume.

\begin{figure*}[!h]
\centering
\includegraphics[width=\textwidth]{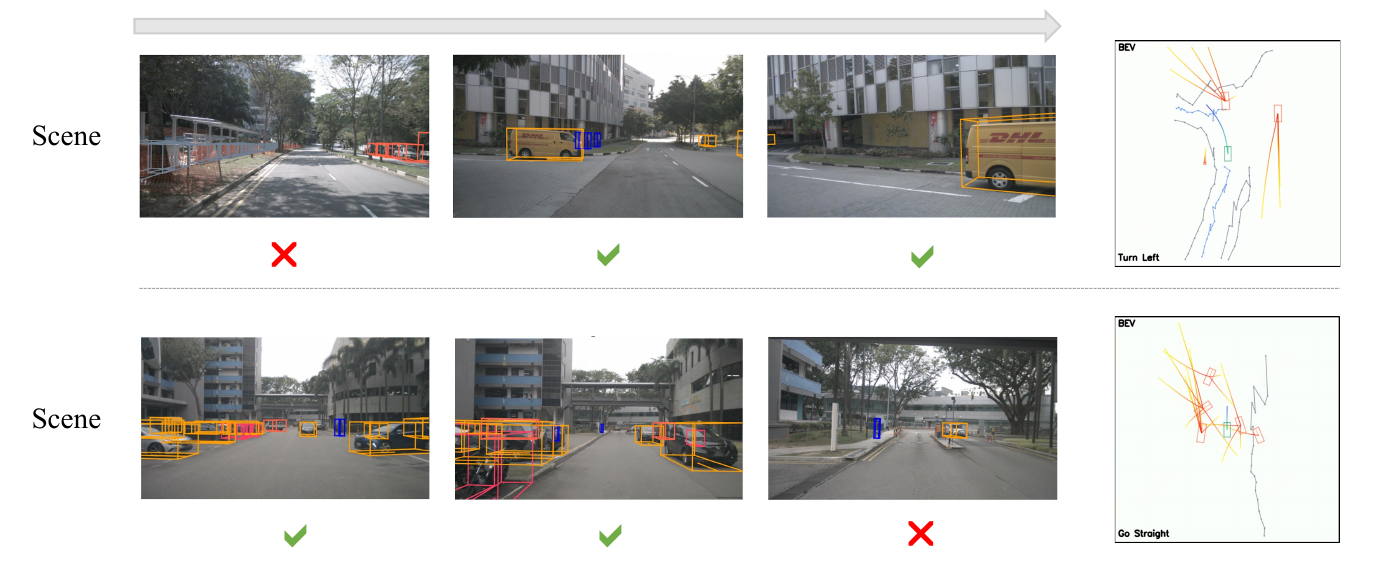} 
\caption{\textbf{Visualization of Selected Scenes and Frames}. The front camera view is used to display the selected data. The \(Scene \) represents a chosen scenario, with \checkmark indicating the extracted key frames. BEV representation is provided to visualize the scenario value more intuitively.}
\label{fig:4}
\end{figure*}

\subsubsection{Visualization of Selected Scenes and Frames.}
The model trained on the initial dataset is used to infer on the remaining data pool. Based on predefined criteria, valuable scenes and key consecutive frames within them are selected. Some results are shown in Fig. \ref{fig:4}.


\begin{figure}[!h]
\centering
\includegraphics[width=\columnwidth]{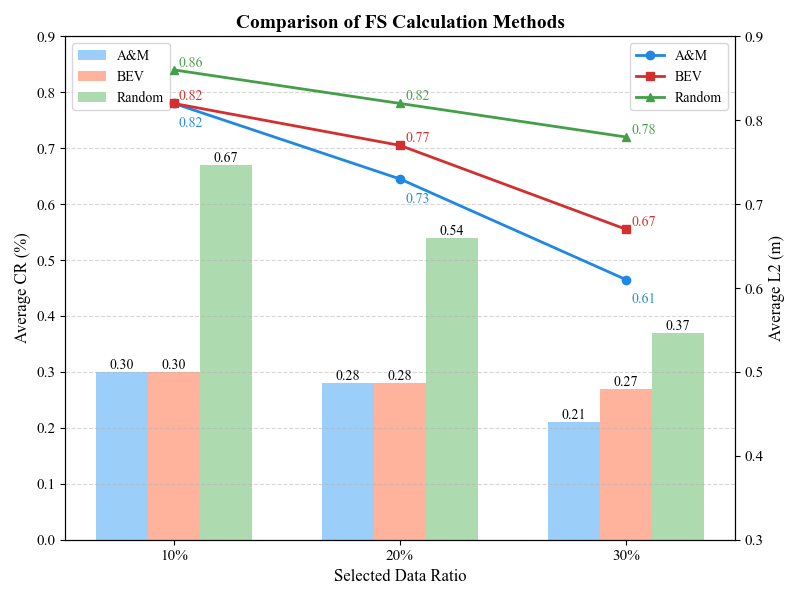} 
\caption{\textbf{Calcutation of $FS$.} We evaluated the impact of different $FS$ calculation methods on the results. Specifically, \( BEV \) represents \( FS \) calculated directly from BEV features, while \( A\&M \) corresponds to calculations based on agent and map features.}
\label{fig:5}
\end{figure}

\subsubsection{Extension of the FS Calculation}
When calculating \( FS \), we begin by converting the \( FS \) of the entire scene to that of key elements. To further explore the robustness and versatility of this approach, we also compare it with alternative forms of \( FS \) calculation. The results, presented in Fig. \ref{fig:5}, reveal that different \( FS \) calculation methods impact the final outcomes, though all demonstrate improved performance compared to random selection. This underscores the potential of data selection based on BEV features and the adaptability of this approach.

\section{Conclusion}
In this paper, we introduce SEAD, a novel active learning strategy specifically designed for E2E-AD. This approach aims to tackle two significant challenges in autonomous driving systems: the high costs associated with data labeling and the issue of long-tail distribution in driving data. Our approach selects initial data based on driving and scene diversity, while incremental data selection is guided by elemental feature offsets in BEV representations. Experimental results indicate that we achieve open-loop performance comparable to training on the entire dataset but with less than 30\% of the labeling budget. Further ablation studies validate the effectiveness of each module within the framework. This work is a preliminary exploration into assessing data value from the BEV perspective, and we hope it will serve as a foundation for future research in this area.

\FloatBarrier
\bibliographystyle{named}
\bibliography{ijcai25}
\clearpage
\end{document}